# DELMAR: Deep Linear Matrix Approximately Reconstruction to Extract Hierarchical Functional Connectivity in the Human Brain

**Wei Zhang**

University of California San Francisco

wei.zhang4@ucsf.edu

**Yu Bao**

James Madison University

bao2yx@jmu.edu

## Abstract

The Matrix Decomposition techniques have been a vital computational approach to analyzing the hierarchy of functional connectivity in the human brain. However, there are still four shortcomings of these methodologies: 1). Large training samples; 2). Manually tuning hyperparameters; 3). Time-consuming and require extensive computational source; 4). It cannot guarantee convergence to a unique fixed point.

Therefore, we propose a novel deep matrix factorization technique called Deep Linear Matrix Approximate Reconstruction (DELMAR) to bridge the abovementioned gaps. The advantages of the proposed method are: at first, proposed DELMAR can estimate the important hyperparameters automatically; furthermore, DELMAR employs the matrix backpropagation to reduce the potential accumulative errors; finally, an orthogonal projection is introduced to update all variables of DELMAR rather than directly calculating the inverse matrices.

The validation experiments of three peer methods and DELMAR using real functional MRI signal of the human brain demonstrates that our proposed method can efficiently identify the spatial feature in fMRI signal even faster and more accurately than other peer methods. Moreover, the theoretical analyses indicate that DELMAR can converge to the unique fixed point and even enable the accurate approximation of original input as DNNs.

## 1 Introduction

Over the last two decades, a variety of computational approaches have been proposed to reconstruct functional architectures of the human brain, e.g., General Linear Modeling (GLM), Graph Theory, Independent Component Analysis (ICA), Sparse Dictionary Learning (SDL), and Low-Rank Matrix Fitting (LMaFit) [1-7]. Nevertheless, most of these methods using a strategy of matrix decomposition are based on a single layer, i.e., the '*shallow*' architecture that is challenging to reconstruct the unsupervised/data-driven fashion of hierarchical and spatially related organization of functional connectivity (FC) in the human brain using resting-state fMRI or task-based fMRI signals [7-10]. The hierarchical reconstruction of spatial FC has been proved by the number of features in *shallow* linear models, e.g., from low to high numbers of independent components in ICA [11, 12], and even realizing that minor networks at the more granular factorization attempt to merge or otherwise recombine to construct more extensive networks at the coarser decomposition. However, it is difficult to reconstruct higher-level FCs since there has been no principled and unsupervised way to reconstruct these hierarchical architectures via *shallow* methods.

Fortunately, with the development of deep learning methods, there have been reported a variety of deep learning models provide an opportunity to reconstruct hierarchical networks, e.g., the Deep Convolutional Auto Encoder (DCAE), Deep Belief Network (DBN), and Convolutional Neural Network (CNN) [9, 10, 13-20]. In addition, the Restricted Boltzmann Machine (RBM) can also be available to extract hierarchical temporal features from fMRI and reconstruct FC networks with impressive accuracy [7, 8]. Moreover, other recent research works have reported the reasonable hierarchical temporal organization of task-based fMRI time series, each with corresponding task-evoked FCs [7-10], using DCAE, RBM, and DBN. In general, these machine learning techniques are thought of as deep nonlinear models, e.g., deep neural networks (DNNs), which are constructed with nonlinear activation functions. Although these nonlinear models such as DBN have recently proven efficacy in the hierarchical spatiotemporal reconstruction of task-evoked fMRI data [7, 9], there are several disadvantages to overcome: 1) large training samples [21-23]; 2) extensive computational resources, e.g., graphics processing units (GPUs) or tensor processing units (TPUs) [7, 9, 13, 23]; 3) manual tuning of hyperparameters, e.g., the number of layers and size of dictionary; some parameters such as sparse trade-off and step-length, are not denoted as hyperparameters [10, 24]; 4) time-consuming training process [16]; 5) non-convergence to the unique fixed point [14, 16, 17]; and 6) "black box" results that are not clear to be explained [16].

**Contribution**. Therefore, we propose a novel deep linear model named Deep Linear Matrix Approximately Reconstruction to overcome the aforementioned shortcomings possessed by the DNNs/Deep nonlinear models. Especially, DELAMAR is superior to other models in the following areas.

*Accurate Approximation*. The reconstruction performance of DELMAR is comparable to DNNs [27, 28]. DELMAR can be treated as polynomials to approximate the original function [29]. In contrast, the DNNs include the nonlinear functions, i.e., activation function, to approximate the original function, which means DNNs can achieve a higher reconstruction accuracy than deep linear matrix factorization. In this work, we prove that the reconstruction of DELMAR is equivalent to DNNs in Theorem 1.1 in section 3, and the proof can be viewed in Appendix A, Supplementary Material.

*Acceleration and Convergence.* Given the assumption of finite dimensionality space and Lemma 1.1 (proof details included in Appendix A), all norms are equivalent and using the Alternation Direction Method of Multiplier (ADMM) [25-28], DELMAR can alternatively optimize all variables and guarantee convergence to the unique fixed point, according to Banach Fixed Point Theorem [30]. Furthermore, the convergence of DELMAR has been proved via Theorem 1.2, and its proof is included in Appendix A. Meanwhile, Corollary 1.1 and 1.2 in Appendix A provide the guidelines to design a deep method: If a given deep model is denoted as a composition of operators and the deep model will converge to a fixed point, it must contain one contraction operator at least, and other operators should be bounded.

*Matrix Backpropagation*. Considering the accumulative error results in poor reconstruction, we utilize a technique of matrix backpropagation (MBP) [31, 32] in order to increase the reconstruction accuracy by layers. MBP implementation can be referred to as Algorithm 5, Appendix D, Supplementary Material.

*Automatic Hyperparameters Tuning*. All hyperparameters of DELMAR, e.g., number of layers and size of components, number of units/neurons in each layer, can be automatically determined via employing the rank reduction operator (RRO) [25, 26]. In contrast, the hyperparameters tuning of DNNs regularly requires other techniques of neural architecture search (NAS), e.g., reinforcement learning, evolutionary optimization, and sampling methods [10]. The details of RRO implementation can be viewed as Algorithm 2, 3, and 4, Appendix D, in Supplementary Material.

From a clinical perspective, DELMAR can also be applied at the individual level, fast even on conventional central processing units (CPUs), compared with DNNs. All theoretical analyses/proofs can be viewed in Appendix A to C included in Supplementary Material. Finally, from an fMRI research perspective, it is vital that DELMAR reconstruct FCs using data from very few experimental individuals compared to DNNs and may prove unparalleled clinical application and efficiency as fMRI's spatial and temporal resolution continues to improve, detecting more of the hierarchy of brain organization.

**Related Works and Methodological Validation.** For reconstruction of FCs, some peer deep linear models, Deep Sparse Dictionary Learning (Deep SDL) [33], Deep Non-negative Matrix Factorization (Deep NMF) [31, 32], and a compositional approach to develop stacked multilayer Independent Component Analysis [34], has been proposed recently. These related deep linear models will be included in methodological validations. To validate the reconstruction performance of DELMAR is divided into two folds: at first, we theoretically analyze the properties of these models in order to predict differences in, e.g., network sparsity, connectivity strength, and convergence velocity; details of theoretical analyses are included in Appendix C. Furthermore, we conduct *in silico* connectivity reconstruction experiments using real fMRI signal time series to validate the predictions of the theoretical analyses for the relative performance of the DELMAR for the reconstruction of FCs. These results lead to clear conclusions about the strengths of DELMAR and provide a guide to the research community for applying this novel reconstruction methodology as well as for developing superb ones.

## 2　Method

This section presents the details of the proposed DELMAR introduced in the last section, including the optimization function of DELMAR, optimizer, the strategy of rank reduction, matrix backpropagation, and accelerated iterative format using orthogonal decomposition.

### 2.1　Deep Linear Matrix Approximation for Reconstruction

DELMAR aims to detect the hierarchical and overlapping organization of FCs more efficiently than previously described data-driven linear reconstruction methods, e.g., ICA, SDL, and LMaFit, [2, 5, 9, 21, 23]. Due to the constraints of spatial independence in ICA, some investigators have reported that ICA cannot easily identify extensively overlapped functional brain networks [2, 6]. Although SDL can efficiently derive spatial features, i.e., functional brain networks, based on resting-state fMRI and task-based fMRI, it is challenging to leverage the dictionary size, sparsity trade-off, and even the number of layers to implement a deep SDL. Specifically, one must heuristically estimate the dictionary size and number of layers. Simply utilizing the same size of dictionary and number of layers can easily vanish the spatial features of deeper layers due to iteratively using the $\ell_1$ norm. Recent deep nonlinear models, such as DBN can successfully reveal the architecture of hierarchical spatiotemporal features. Unfortunately, the probabilistic energy-based model of DBNs necessarily requires a large number of training samples to avoid overfitting. DELMAR excels in the deep models because it does not require any manual hyperparameter tuning to decompose the input signal matrix. Furthermore, since DELMAR is a deep linear model, it enables faster detection of latent features than DBN while only requiring conventional CPUs and an efficient non-fully connected architectures of DELMAR without activation functions is easy to be optimized.

In short, DELMAR can be approximately considered as a deep linear matrix factorization with the additional techniques to automatically determine all crucial hyperparameters and accelerate the updates of all variables.

The optimization function governing DELMAR taking $k^{\text{th}}$ layers as an example as:

$$min_{Z_i \in \mathbb{R}^{m \times n}} \bigcup_{i=1}^{k} \|Z_i\|_1 \quad (1\text{-}1)$$

$$s.t. \prod_{i=1}^{k} X_i Y_k + Z_k = S \quad (1\text{-}2)$$

$$X_i Y_i \leftarrow \mathcal{R}(Y_{i-1}), \forall\ 2 \leq i \leq k \quad (1\text{-}3)$$

where $\{X_i\}_{i=1}^{k}$ represents the hierarchical weight/mixing matrices, e.g., $X_i$ is the weight/mixing of the $i^{\text{th}}$ layer. Furthermore, $\{X_i\}_{i=1}^{k}$ is also treated as the time series in GLM and the weight matrix in ICA and DBN, and $M$ defines the total number of layers. Similarly, $\{Y_i\}_{i=1}^{k}$ represents the hierarchical spatial features, e.g., $Y_i$ is the FCs of the $i^{th}$ layer. And, $\{Z_i\}_{i=1}^{k}$ are the matrices of background components, which is usually reconstructed as the noise components due to the sparsity of these matrices. Finally, $\mathcal{R}$ represents the operator RRO to automatically estimate the hyperparameters and more details can be found in the following section and Appendix D.

Naturally, like multiple layers stacked model, e.g., DBN, we assume the spatial features $Y_{i-1}$ can be decomposed as a deeper dictionary $X_i$ and spatial features $Y_i$. Therefore, the original input data $S$ can be decomposed as $\prod_{i=1}^{k} X_i Y_k + Z_k$.

Before optimizing Eq. (1), it is required to convert Eq. (1) into an augmented Lagrangian function. In addition, in Eq. (2), $\beta > 1$ denotes a penalty parameter introduced in the following Augmented Lagrangian Function. Intuitively, to avoid manually tuning the spare trade-off, the sparse trade-off of $\bigcup_{i=1}^{k} \|Z_i\|_1$ is defined as $\frac{1}{\beta}$ to control the sparsity levels of background components.

The Augmented Lagrangian Function for $k^{th}$ layer can be expressed as:

$$\mathcal{L}_\beta(\prod_{i=1}^{k} X_i, Y_k, Z_k, e_k) \stackrel{\text{def}}{=} \frac{\beta}{2}\left\|\prod_{i=1}^{k} X_i Y_k - S\right\|_F^2 + \langle \prod_{i=1}^{k} X_i Y_k - S, e_k \rangle + \frac{1}{\beta}\|Z_k\|_1 \quad (2)$$

Since the Augmented Lagrangian Function $\mathcal{L}_\beta(\prod_{i=1}^{k} X_i, Y_k, Z_k, e_k)$ in Eq. (2) is not jointly convex, a natural idea is to employ alternating optimization that only a single variable is updated in each iteration while other variables are treated as constants [27, 28]. Therefore, an ADMM optimizer could be used to implement alternative optimization [25-28] with $e_k$ in Eq. (2) representing the multiplier of ADMM. Moreover, the $\ell_1$ norm of $Z_k$ shown in Eq. (2) can be solved directly using the shrinkage method [35]. Since the convergence of ADMM has been proved as geometric and linear convergence [27, 28], we prove that the ADMM optimizer can guarantee the convergence of DELMAR to the unique fixed point, based on some prerequisites, such as assumption of finite dimensionality space. The detailed proof can be viewed in Theorem 1.2, Appendix A.

The iterative format of ADMM to optimize Eq. (2) can be organized as follows:

$$X_k^{it+1} \leftarrow argmin_{X_k^{it+1} \in \mathbb{R}^{m \times h_k}} \mathcal{L}_\beta(X_k^{it}, Y_k^{it}, Z_k^{it}, e_k^{it}) \quad (3\text{-}1)$$

$$Y_k^{it+1} \leftarrow argmin_{Y_k^{it+1} \in \mathbb{R}^{h_k \times n}} \mathcal{L}_\beta(X_k^{it+1}, Y_k^{it}, Z_k^{it}, e_k^{it}) \quad (3\text{-}2)$$

$$Z_k^{it+1} \leftarrow argmin_{Z_k^{it+1} \in \mathbb{R}^{m \times n}} \mathcal{L}_\beta(X_k^{it+1}, Y_k^{it+1}, Z_k^{it}, e_k^{it}) \quad (3\text{-}3)$$

$$e_k^{it+1} \leftarrow e_k^{it} + \eta\beta \cdot (\prod_{i=1}^{k} X_i Y_k + Z_k^{it+1} - S) \quad (3\text{-}4)$$

As mentioned earlier in Eq. (2), there are three parameters, $Y_k^{it}, Z_k^{it}$ and $e_k^{it}$ to be optimized with detailed process illustated by Eq. (3-1) to Eq. (3-4). Specifically, Eq. (3-1) shows the process of updating $X_k^{it}$ where $X_k^{it+1}$ equals the minimum of $\mathcal{L}_\beta(X_k^{it}, Y_k^{it}, Z_k^{it}, e_k^{it})$ while $Y_k^{it}, Z_k^{it}$ and $e_k^{it}$ (the current iteration is represented as *it*) are treated as constants. Similarly, Eqs. (3-2) and (3-3) presents the update process for $Y_k^{it}$ and $Z_k^{it}$ respectively. Finally, in Eq. (3-4), we update the multiplier based on the current error. In addition, in Eq. (3-4), $\eta$ denotes the step length where $\eta > 1$ represents the relaxation optimization [25, 26, 28].

## 2.2 Rank Reduction Operator (RRO) for Automatic Hyperparameters Tuning

One of the most critical contributions of DELMAR is the automatic estimation of all hyperparameters by employing RRO to reduce the input matrix's dimension progressively. In detail, DELMAR can continuously decompose the input matrix for each layer until the rank of the decomposed matrix is equal to 1. In addition, this rank estimator RRO employs a technique of rank-revealing by constantly using orthogonal decomposition, in this case via *QR* factorization [25, 26]. Obviously, the advantage of *QR* is that it is faster than SVD with fewer requirements of the input matrix. For instance, it is challenging for SVD to decompose a sparse matrix [25, 26].

Assume $r^*$ is the initial rank estimation of the input matrix $S$ and $r$ is the optimal rank estimation of $S$; if $r^* \geq r$ holds, the diagonal line of the upper-triangular matrix in the $QR$ factorization can be performed using the input matrix $S$. Then, similarly, $S$ is replaced by $Y_k, k \in [1, M]$ and $M$ assumes the total number of layers, to perform further factorization. The definition of RRO is shown below:

$$\mathcal{R}\begin{bmatrix} a_1 \\ a_2 \\ \vdots \\ a_{n-1} \\ a_n \end{bmatrix} = \begin{bmatrix} a_1^{(1)} \\ a_2^{(1)} \\ \vdots \\ a_{n-2}^{(1)} \\ a_{n-1}^{(1)} \end{bmatrix} \mathcal{R}^k \begin{bmatrix} a_1 \\ a_2 \\ \vdots \\ a_{n-1} \\ a_n \end{bmatrix} = \begin{bmatrix} a_1^{(1)} \\ a_2^{(1)} \\ \vdots \\ a_{n-k-1}^{(1)} \\ a_{n-k}^{(1)} \end{bmatrix} = [\hat{a}] \tag{4}$$

where $\mathcal{R}$ represents the RRO operator, $\mathcal{R}^k$ denotes the combination of $\mathcal{R}$ as $k$ times, such as $\mathcal{R}^2 \stackrel{\text{def}}{=} \mathcal{R} \cdot \mathcal{R}$, the input matrix is denoted as $[a_1, a_2, \cdots, a_n]$, and we have $r^* = rank([\hat{a}]) < r = rank(\mathcal{R}^k[a_1, a_2, \cdots, a_n])$. Also, if $k$ is large enough, we have $rank([\hat{a}]) = 1$. Furthermore, Theorem 2.4 shows that $\mathcal{R}$ is a bounded operator, such as $\mathcal{R}: \mathbb{R}^{S \times T} \to \mathbb{R}^{S \times T}$, $\|\mathcal{R}\| < \infty$ where the prove can by found in Appendix B, Supplementary Material.

Moreover, we employ the weight ratio (*WR*) and weight difference (*WD*) to generate a new vector based on the diagonal elements of matrix $R$. The maximum value of *WR* and *WD* are then utilized to determine the rank of the input matrix. The details of pseudocode can be founded in Algorithm 3 and 4, in Appendix D, Supplementary Material. In the following equations, assume $T > S$ and $T$ represents the total number of columns, for $i \in [1, T]$, we have:

$$\begin{aligned} d_i &\leftarrow |R_{ii}| \\ wr_i &\leftarrow \frac{d_i}{d_{i+1}} \end{aligned} \tag{5}$$

where $d_i$ represents a diagonal element $R_{ii}$ of matrix $R$ derived by $QR$ decomposition and $wr_i$ represents the $i$th value of *WR* calculated by the ratio of the current diagonal element and the next diagonal element of $R$.

Furthermore, *WD* is calculated as:

$$wd_i \leftarrow \frac{|d_i - d_{i-1}|}{\sum_{j=1}^{i-1} d_j} \tag{6}$$

*WD* is the absolute difference between the current diagonal element and the previous one divided by the cumulative sum of all previous diagonal elements. Since *WD* and *WR* are the cumulative difference and ratio, respectively, the matrix dimension can be reduced by one.

The details of RRO implementation can be viewed in Algorithms 2-4, included in Appendix D. It demonstrates that RRO can continuously reduce the dimensions of the original data and comparably retain the vital components as sparse Principal Component Analysis (sPCA) [27]. By continuously using the technique of rank reduction, DELMAR can automatically obtain the weight matrix size and the number of layers.

### 2.3   Matrix Backpropagation to Reduce the Accumulative Error

Another important technique named matrix backpropagation is applied to DELMAR in order to reduce the potential accumulative error. After updating all variables, if assume the total layers number is $M$, Eq. (7) describes the variables after updates as $\{X_i\}_{i=1}^M$ and $\{Y_i\}_{i=1}^M$ before utilizing the matrix backpropagation when decomposition is finished [31], [32], [42].

$$\begin{aligned} \hat{Y}_k &\leftarrow Y_k, k = M \\ \hat{Y}_k &\leftarrow X_{k+1} \hat{Y}_{k+1}, k < M \end{aligned} \tag{7}$$

Meanwhile, we denote the production of hierarchical weight matrices as $\psi$ in Eq. (8)

$$\psi \leftarrow \prod_{i=1}^{k-1} X_i \tag{8}$$

Then, the following equation describes the vital step of matrix propagation to update variables of $\{Y_i\}_{i=1}^M$, and $\{Z_i\}_{i=1}^M$ represents the hierarchical background noise.

$$\hat{Y}_k^+ \leftarrow Y_k^+ \odot \sqrt{\frac{[\psi^T Z_k]^+ + [\psi^T \psi]^- \hat{Y}_k^+}{[\psi^T Z_k]^- + [\psi^T \psi]^+ \hat{Y}_k^+}}; \tag{9}$$

More details can be viewed in Algorithm 5, Appendix D.

Moreover, because the calculation of matrix pseudo-inverse is time-consuming [25], [26], we are inspired to employ the orthogonal projection to calculate the transpose of the matrix instead of the pseudo inverse.

$$QQ^T = I \Longrightarrow Q^{-1} = Q^T \tag{10}$$

Based on Augmented Lagrangian Function in Eq. (2) and orthogonal projection introduced in [41], we provide the accelerated iterative format as:

$$X_{k+1} \leftarrow QR\left(\left(S - \frac{e_k}{\beta}\right) Y_k^T\right) \tag{11}$$

Furthermore, for $\{Y_i\}_{i=1}^M$, we have:

$$Y_{k+1} \leftarrow QR(X_k^T \left(S - \frac{e_k}{\beta}\right)) \tag{12}$$

Using Eqs. (11) and (12) to update variables $\{X_i\}_{i=1}^M$ and $\{Y_i\}_{i=1}^M$ are more efficient than using the iterative format in Eqs. (3-1) and (3-2). The derivation details can be viewed in the last section, Appendix A, Supplementary Material.

## 3 Approximation, Acceleration, and Convergence of DELMAR

Another vital contribution of DELMAR is the accurate approximation to the original input which yields competitive reconstruction accuracy compared to other deep linear models, such as DNNs. This section presents the theoretical analysis of the approximation and convergence properties for DELMAR.

Since DELMAR can be treated as a composition of polynomials, the following theorem demonstrates that DELMAR can approximate any real functions even with very few values approaching infinite by only requiring enough layers. Therefore, DELMAR can guarantee the comparable reconstruction accuracy to DNNs that employ the nonlinear activation function which often achieve a higher reconstruction accuracy. The proof of Theorem 1.1 can be viewed in Appendix A, Supplementary Material.

At first, we introduce the definition of matrices polynomials as: $\forall k \in \mathbb{N}, P_{2k}(X) = (XX^T)^k$, $P_{2k+1}(X) = (XX^T)^k X, X \in \mathbb{R}^{S \times T}$; and $\{P_n(X)\}_{n=1}^N$ denotes a series of matrix polynomials, for example: $\{P_n(X)\}_{n=1}^3 = \{X, XX^T, XX^T X\}, X \in \mathbb{R}^{S \times T}$; moreover, $\{P_n\}_{n=1}^\infty$ denoted on $\mathbb{R}^{S \times T}$ can be easily proved as a polynomial ring ($\{P_n\}_{n=1}^\infty, +, \times$) [36, 37]. Then we introduce the Theorem 1.1 to describe the superiority of DELMAR for the high accuracy of approximating the original input matrix.

***Theorem 1.1*** **(Approximation Superiority of DELMAR)** Given a real function $f: \mathbb{R}^{S \times T} \to \mathbb{R}^{S \times T} \cup \{\pm \infty\}$ and $m(\{X \in \mathbb{R}^{S \times T}: f(X) = \pm \infty\}) = 0$, where $m(\cdot)$ represents the Lebesgue measure [29]. For $X \in \mathbb{R}^{S \times T}$ and the series of matrix polynomials $\{P_n(X)\}_{n=1}^N$, we have: Given $\forall \varepsilon > 0$, there exists $N_0$ such that for any $N > N_0$, $\|\{P_n(X)\}_{n=1}^N - f(X)\| \leq \varepsilon$; when $N \to \infty$, we have: $\lim_{N \to \infty} \{P_n(X)\}_{n=1}^N = f(X)$; however, for any *shallow* model, since $N$ is bounded, we only have: $\|\{P_n(X)\}_{n=1}^N - f(X)\| \leq M$ where $M$ is a small constant.

The following lemmas and corollaries provide a guideline to construct a deep model that can converge to a fixed point. As discussed before, if a computational model can be treated as a composition of operators, the convergence to a fixed point requires that at least one operator are compact and the others are bounded. The fundamental assumption in this work only considers the finite-dimensional space that guarantees the equivalent of norms discussed in this work (please see Lemma 1.1, Appendix A, in Supplementary Material).

***Lemma 1.2*** **(Contraction of Operators Combination)** Given two contraction mappings $\Phi_1$ and $\Phi_2$, the composite mapping $\Phi_2 \cdot \Phi_1$ must be contractive.

***Corollary 1.1*** **(General Contraction Operator/Guideline to Design a Deep Model)** Suppose $\{\Phi_i\}_{i=1}^K$, $\forall \Phi_i\ i \in \mathbb{N}$, $\Phi_i: \mathbb{R}^{S \times T} \to \mathbb{R}^{S \times T}$ is a series of operators. According to Lemma 1.2, for any combination of operators: $\Phi_K \cdot \cdots \cdot \Phi_2 \cdot \Phi_1$, if at least a single operator $\Phi_i$ is a contraction operator and the other operators are bounded, such as $\forall i \neq k\ \|\Phi_i\| \leq M$, the combination of operator series $\Phi_K \cdot \cdots \cdot \Phi_2 \cdot \Phi_1$ is a contraction operator if and only if $\prod_{i=1}^K \|\Phi_i\| < 1$ or $\|\Phi_j\| < 1, j \in \mathbb{N}, 1 < j < K$.

***Corollary 1.2*** **(Iterative Contraction Operator)** Suppose $\{\Phi_i\}_{i=1}^K$, $\forall \Phi_i\ i \in \mathbb{N}$, $\Phi_i: \mathbb{R}^{S \times T} \to \mathbb{R}^{S \times T}$ is a series of operators. According to Lemma 1.2, for any combination of operators: $\Phi_K \cdot \cdots \cdot \Phi_2 \cdot \Phi_1$, if at least a single operator $\Phi_i$ is a contraction operator, and the other operators are bounded, such as $\forall i \neq k, \|\Phi_i\| \leq M$, the combination of operator series $\Phi_K^n \cdot \cdots \cdot \Phi_2^n \cdot \Phi_1^n$ is a contract operator if and only if $\lim_{n \to \infty} \prod_{i=1}^K \|\Phi_i\|^n = c < 1$.

Since optimizer ADMM can be proved as a compact operator and other operations of DELMAR are bounded, e.g., random initialization, sparse operation in DELMAR can guarantee the convergence to a fixed point, according to Banach Theorem. All theoretical analyses can be viewed in Appendix A, B and C, Supplementary Material.

Finally, Theorem 1.2 below demonstrates that a computational model with multiple variables satisfying Corollary 1.2 can converge to a fixed point via an alternative strategy. The proof of Theorem 1.2 can be viewed in Appendix A, Supplementary Material.

***Theorem 1.2*** **(Alternative Convergence of DELMAR)** Suppose $\{\mathcal{F}_{i,j}\}_{i,j=1}^\infty$, $\mathcal{F}_{i,j}: \mathbb{R}^{M \times N} \to \mathbb{R}^{M \times N}$ and $\{\mathcal{H}_j\}_{j=1}^\infty$, $\mathcal{H}_j: \mathbb{R}^{M \times N} \to \mathbb{R}^{M \times N}$ are two seriers of continuous operators defined on a finite dimensional space. If we have: $\lim_{i \to \infty} \mathcal{F}_{i,j} \to \mathcal{H}_{M,j}$ and $\lim_{j \to \infty} \mathcal{H}_{M,j} \to \mathcal{G}$, then, $\exists \lim_{k \to \infty} \mathcal{F}_{i_k, j_k} \to \mathcal{G}$ holds.

## 4    Results

We employ all subjects' resting-state fMRI signals from healthy individuals retrieved from the Consortium for Neuropsychiatric Phenomics (CNP) (https://openfmri.org/dataset/ds000030/) in order to validate the reconstruction performance of DELMAR. To avoid heterogeneous parameter tuning, all hyperparameters are tuned by following the procedure of the hyperparameters estimations of DELMAR. The estimated layer for DELMAR including all subjects is 2 with the size of the first and the second layers equal to 25 and 6, respectively [12]. Other parameters are tuned in [31]-[34]. All abbreviations of templates can be referred in Table S1, Appendix A, Supplementary Material.

Figure 1 and 2 show a reconstruction of the eight FCs at 1st layer via DELMAR and other peer methods, e.g., Deep SDL, Deep FICA, and Deep NMF, compared with templates [12]. The qualitative validations demonstrate that the reconstruction of first layer FCs of DELMAR and Deep SDL is comparable to original templates, especially considering the intensity.

Based on the simple observation of Figures 1 and 2, the DELMAR reconstruction is consistent with the original template/ground truth networks shown in the first column [12]. Furthermore, the intensity of reconstructed FCs is very similar to the templates. The theoretical explanation of the highest intensity matching of DELMAR is included in Appendix C. All representative slices of reconstructed FCs can be viewed via Figure S1, Appendix E, Supplementary Material.

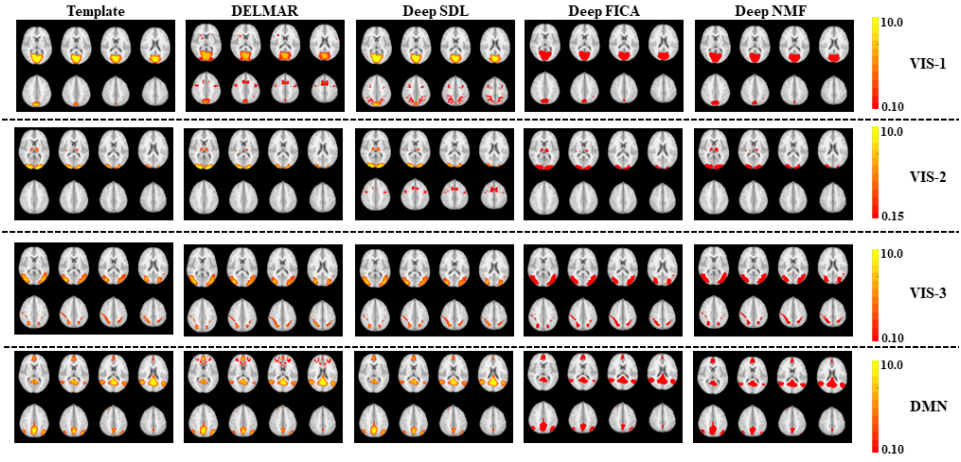

Figure 1. This figure presents eight representative slices of the reconstructed four FCs via DELMAR and other three peer methods.

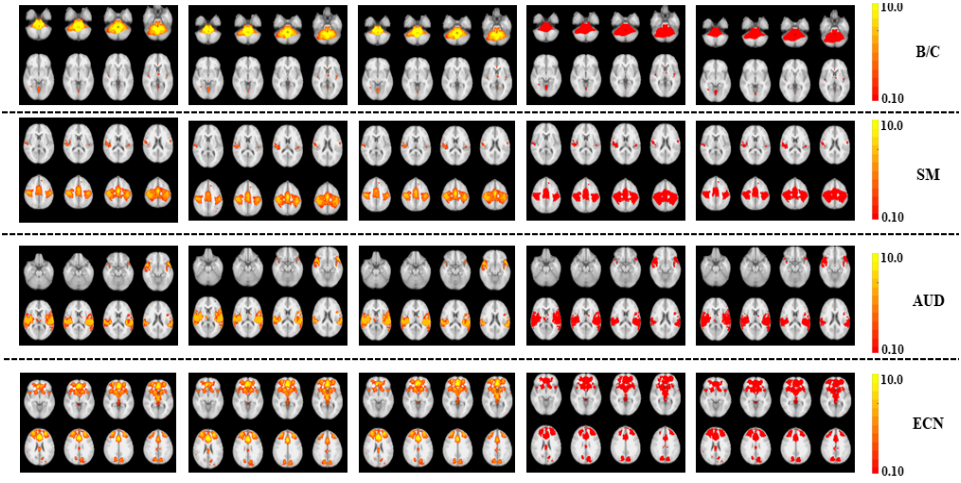

Figure 2. This figure presents eight representative slices of another four reconstructed FCs via DELMAR and other three peer methods.

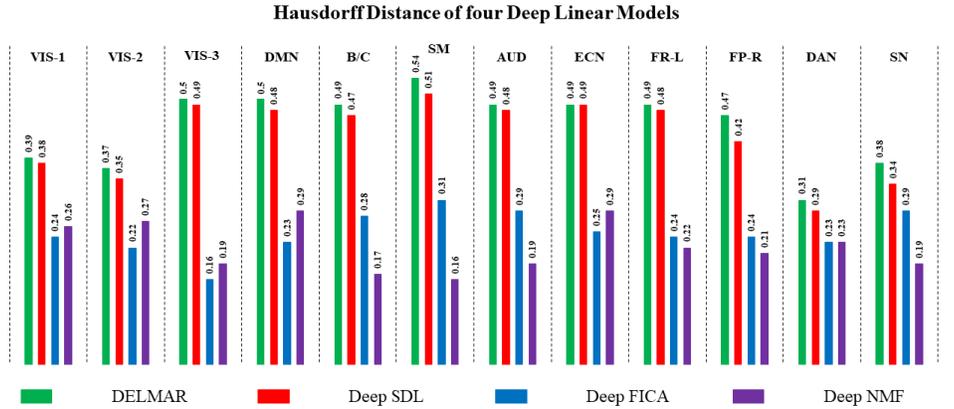

Figure 3. All quantitative similarity analytics of the reconstructed twelve 1st layer FCs and the twelve ground truth templates [12]. Green bars demonstrate that DELMAR can reconstruct twelve FCs with a very high similarity, compared with the ground truth templates. All details of the ground truth templates can be viewed in Table S1, Appendix A, Supplementary Material.

The qualitative results show that the reconstruction of FCs at the first layer derived via DELMAR are better than the other three peer deep linear algorithms, that is, the reconstructed FCs of DELMAR provide more similar intensity and spatial similarity with the ground truth templates, using Hausdorff distance [6].

Finally, to examine the reproducibility of DELMAR and other three peer algorithms, we randomly separate the original input data into two independent sets as $FC_{test}$ and $FC_{retest}$ shown in Eq. (13). The quantitative results of reproducibility are presented in Figure 5(c).

$$fc_i \in FC_{test}, fc_j \in FC_{retest}, FC_{test} \cap FC_{retest} = \emptyset \qquad (13)$$

$$reproduceVal \leftarrow corr(fc_i, fc_j)$$

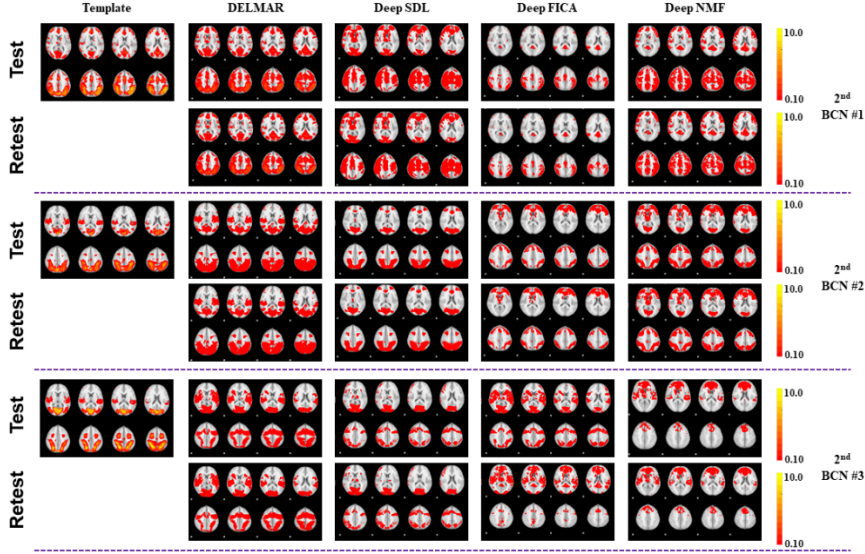

Figure 4. All quantitative similarity analyses of the reconstructed twelve 1$^{st}$ layer FCs and the 12 ground truth templates. The blue bars demonstrate that DELMAR can reconstruct twelve FCs with a very high similarity, compared with the ground truth templates. All details of the ground truth templates can be viewed in Table S1, Appendix A, Supplementary Material.

Furthermore, the quantitative comparison of the first layer FCs are provided in the Figures 3, 4, and the supplementary materials. Figure 3 compares the reconstructed first layer FCs of DELMAR with those for the other three peer methods based on the original templates. The similarity is derived based on Hausdorff Distance [9, 10]. We also provide theoretical analyses to explain why DELMAR and Deep SDL can achieve a superb intensity matching (please see columns first and second). In short, the intensity of the reconstructed FCs/features depends on the norm of operators, i.e., each algorithm is denoted as a composition of several operators. More details can be found in Definition 2.1 to 2.4, Theorem 2.1 to 2.10, Appendix B and Theorem 3.1 and 3.2, Appendix C, Supplementary Material.

Moreover, Figure 5 presents the reconstruction of the 2nd FCs of DELMAR and the other three peer methods. Generally speaking, these linear models of matrix approximation can successfully reconstruct hierarchical FCs. These higher-level FCs can be considered as the recombination of *shallow* FCs. Obviously, DELMAR can reconstruct the most recombined FCs at the second layer. Although Deep SDL can reconstruct six 2nd layer FCs, the last two FCs (in columns fifth and sixth) overlap extensively with *shallow* templates, e.g., B/C and FP-L, respectively.

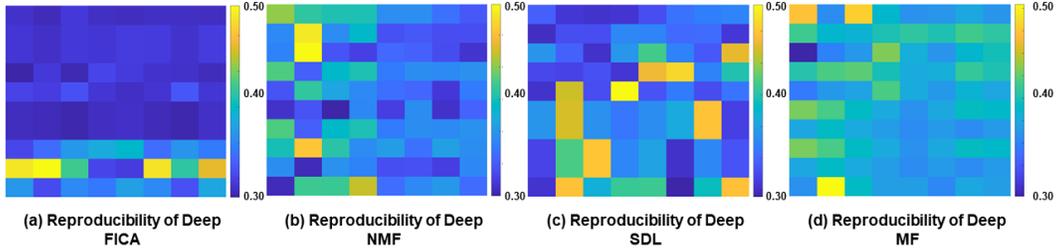

(a) Reproducibility of Deep FICA  (b) Reproducibility of Deep NMF  (c) Reproducibility of Deep SDL  (d) Reproducibility of Deep MF

Figure 5. The reproducibility presentation of the higher-level FCs (2nd layer FCs) demonstrates that DELMAR and Deep SDL can reconstruct more reproducible 2nd layer FCs while Deep FICA and Deep NMF reconstruct less reproducible FCs.

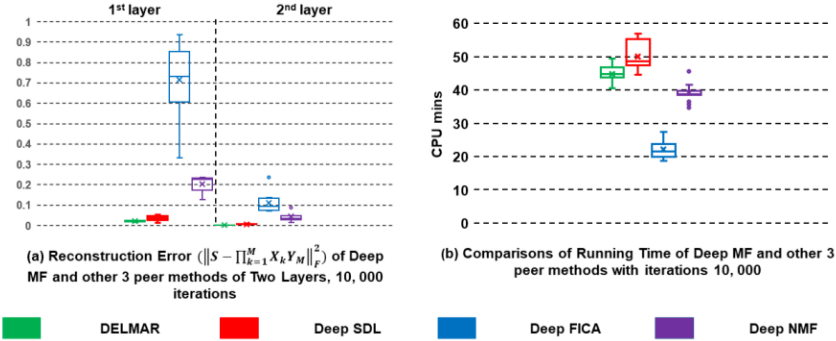

(a) Reconstruction Error ($\|S - \prod_{k=1}^{M} X_k Y_M\|_F^2$) of Deep MF and other 3 peer methods of Two Layers, 10,000 iterations

(b) Comparisons of Running Time of Deep MF and other 3 peer methods with iterations 10,000

DELMAR    Deep SDL    Deep FICA    Deep NMF

Figure 6. This figure demonstrates the reconstruction accuracy of 15 subjects via DELMAR and the other three peer methods. DELMAR (showed in green bar) outperforms the other algorithms with less CPU time.

## 5  Conclusion

In this work, we propose the novel method, DELMAR that not only requires fewer training datasets but also determines all hyperparameters automatically; furthermore, DELMAR detects the hierarchical FCs more efficiently than the other selected peer algorithms due to the acceleration strategy used to update all variables; in addition, DELMAR can be introduced to clinical translational research, since these reconstructions of biomarkers can be further validated as clinically actionable biomarkers for diagnosis, prognosis and treatment monitoring of neurodevelopmental, neurodegenerative, and psychiatric disorders [38-40].

Moreover, this study demonstrates a detailed procedure with strong theoretical support of the implementation of ADMM and parameter estimation for DELMAR. This alternative optimization algorithm is particularly well suited to optimize convex/alternative convex problems [27], [28] and also utilizes RRO for data-driven determination of all hyperparameters, which can be considered an advanced factorization method. This is a major contribution over other peer conventional *shallow*, data-driven fMRI connectivity reconstruction methods presented in this study.

Finally, the theoretical analyses and the fMRI validation procedure provided in this work should synergize further development of deep linear models optimized for real-world biomedical imaging applications, with DELMAR being one of the optimal algorithms for fMRI hierarchical functional connectivity mapping. Overall, we believe that DELMAR can serve as an initial point for fruitful future works with the promising direction of deep matrix factorization techniques.

# Supplemental Material

## Appendix A

Table S1. All abbreviations of template FCs used as the Ground Truth FCs in the following validations.

| Name/Number | Abbreviation | Name | Abbreviation |
|---|---|---|---|
| Primary Visual Network/1 | VIS-1 | Auditory Network/7 | AUD |
| Perception Visual Shape Network/2 | VIS-2 | Executive Control Network/8 | ECN |
| Perception Visual Motion Network/3 | VIS-3 | Left Frontoparietal Network/9 | FP-L |
| Default Mode Network/4 | DMN | Right Frontoparietal Network/10 | FP-R |
| Brainstem & Cerebellum Network/5 | B/C | Dorsal Attention Network/11 | DAN |
| Sensorimotor Network/6 | SM | Salience Network/12 | SN |

Table S2. All definitions of operators and their norms.

| Operator | Definition | Operator/Norm | Definition |
|---|---|---|---|
| $\mathfrak{A}$ | DELMAR | $\mathcal{U}$ | Update Operator of Deep NMF |
| $\mathfrak{N}$ | Deep NMF | $\mathcal{F}$ | Operator of Fixed-Point Algorithm |
| $\mathfrak{L}$ | Deep SDL | $\mathcal{C}$ | Consistent Operator |
| $\mathfrak{T}$ | Deep FICA | $\varphi$ | Norm of ADMM |
| $\mathcal{M}$ | Random Initialization Operator | $\rho$ | Norm of GD |
| $\mathcal{S}$ | Sparsity Operator | $\mu$ | Norm of Fixed-Point Algorithm |
| $\mathcal{P}$ | Principal Component Analysis (PCA) | $\gamma$ | Norm of Normalization in Deep NMF |
| $\mathcal{A}$ | ADMM | $\delta$ | Norm of Updating Deep NMF |
| $\mathcal{G}$ | GD | $C$ | Norm of Input fMRI Matrix |
| $\mathcal{N}$ | Normalization | $S$ | Input fMRI Signal Matrix |
| $\mathcal{R}$ | Rank Reduction Operator | $\mathfrak{C}$ | Set of Consistent Operators |

***Assumption 1.1*** For any operator discussed in this study, we have: $\forall \mathcal{C} \in \mathfrak{C}, \mathcal{C}: \mathbb{R}^{S \times T} \to \mathbb{R}^{S \times T}, S < \infty, T < \infty$. This assumption demonstrates that all operators are mapping from the finite dimensional space to another finite dimensional space, which is also reasonable in the real world.

***Lemma 1.1*** **(Norm Equality)** Given any arbitrary norm $\|\cdot\|$ and/or their finite linear combination $\sum_{i=1}^{n} k_i \|\cdot\|$ denoted based on any finite set, this norm or their finite linear combination is equivalent to $\ell_2$ norm (e.g., $\|\cdot\|_2$).

***Proof***: We denote $\ell_1$ and $\ell_2$ norm in finite dimensions, such as:

$$\|a\|_1 = \sum_{i=1}^{n} |a_i|$$

$$\|a\|_2 = \left(\sum_{i=1}^{n} a_i^2\right)^{\frac{1}{2}} \quad (A.1)$$

$$a = [a_1, a_2, \cdots, a_n]$$

Obviously, since all norms are non-negative, according to Eq. (A.1), we have:

$$\sum_{i=1}^{n} a_i^2 \leq \left(\sum_{i=1}^{n} |a_i|\right)^2 \quad (A.2)$$

Eq. (A.2) implies:

$$\|a\|_2 \leq \|a\|_1 \quad (A.3)$$

And, based on Cauchy-Schwarz inequality, we have:

$$\|a\|_1^2 = \left(\sum_{i=1}^{n} |a_i| \cdot 1\right)^2 \leq \sum_{i=1}^{n} a_i^2 \cdot \sum_{j=1}^{n} 1^2 = \|a\|_2^2 \cdot n \quad (A.4)$$

It implies:

$$\frac{1}{\sqrt{n}} \|a\|_1 \leq \|a\|_2 \tag{A.5}$$

According to the theorem of norm equality [30], given an arbitrary finite dimensional space, if and only if the following inequality holds:

$$c\|\cdot\|_2 \leq \|\cdot\| \leq C\|\cdot\|_2 \tag{A.6}$$

Thus, the norm $\|\cdot\|$ is equivalent to $\|\cdot\|_2$. Since Eq. (B.3) and Eq. (B.4) hold, we have:

$$c\frac{1}{\sqrt{n}} \|a\|_1 \leq \|a\|_2 \leq \|a\|_1 \tag{A.7}$$

It implies $\|\cdot\|_1$ and $\|\cdot\|_2$ are equivalent. Similarly, we can prove $\sum_{i=1}^n k_i \|\cdot\|$ is also equivalent to $\|\cdot\|_2$.

***Theorem 1.1*** **(Superiority of Deep Linear Models)** Given a real function $f: \mathbb{R}^{S \times T} \to \mathbb{R}^{S \times T} \cup \{\pm\infty\}$ and $m(\{X \in \mathbb{R}^{S \times T}: f(X) = \pm\infty\}) = 0$. And $m(\cdot)$ represents the Lebesgue measure. If considering $X \in \mathbb{R}^{S \times T}$ and the series of matrix polynomials $\{P_n(X)\}_{n=1}^N$, we have: if $N$ is large enough, we have: $\forall \varepsilon > 0 \|\{P_n(X)\}_{n=1}^N - f(X)\| \leq \varepsilon$; if $N \to \infty$, we have: $\lim_{N \to \infty} \{P_n(X)\}_{n=1}^N = f(X)$; however, for any shallow model, since $N$ should be bounded, we only have: $\|\{P_n(X)\}_{n=1}^N - f(X)\| \leq M$.

***Proof***: The matrix polynomials are defined as: $\forall k \in \mathbb{N}\ P_{2k}(X) = (XX^T)^k$, $P_{2k+1}(X) = (XX^T)^k X$, $X \in \mathbb{R}^{S \times T}$;

And $\{P_n(X)\}_{n=1}^N$ defines a series of matrix polynomials, for example: $\{P_n(X)\}_{n=1}^3 = \{X, XX^T, XX^T X\}$, $X \in \mathbb{R}^{S \times T}$;

Moreover, it is easy to prove $\{P_n\}_{n=1}^\infty$ denoted on $\mathbb{R}^{S \times T}$ as a ring $(\{P_n\}_{n=1}^\infty, +, \times)$, and it also demonstrates $\{P_n(X)\}_{n=1}^N \in (\{P_n\}_{n=1}^\infty, +, \times)$, e.g., $(\{P_n\}_{n=1}^\infty, +, \times) \supseteq \sum_{i=1}^M \{P_n(X)\}_{n=1}^N$ [36, 37].

According to Лузин (Luzin) Theorem [29], we have a close set:

$$F_n \subset F_{n+1} \subset \cdots \subseteq \mathbb{R}^{S \times T}$$

$$m(\mathbb{R}^{S \times T} \setminus F_k) = \frac{1}{k}, k \in \mathbb{N} \tag{A.8}$$

$$f \in C(F_k)$$

Then we have a consistent real function $g(X)$, and obviously we have:

$$g(X) = f(X) \tag{A.9}$$

Since for any continuous real function, we have:

$$|g(X) - P_k(X)| < \frac{1}{k} \tag{A.10}$$

Let $\mathcal{F} = \bigcup_{k=1}^\infty F_k$, and obviously we have:

$$m(\mathbb{R}^{S\times T}\setminus F_k) = m(\mathbb{R}^{S\times T}\setminus \cup_{k=1}^{\infty} F_k) = \cap_{k=1}^{\infty} m(\mathbb{R}^{S\times T}\setminus F_k) = \cap_{k=1}^{\infty}\frac{1}{k} = 0 \tag{A.11}$$

Moreover, it is easy to prove $\{P_n\}_{n=1}^{\infty}$ denoted on $\mathbb{R}^{S\times T}$ as a ring $(\{P_n\}_{n=1}^{\infty}, +, \times)$, and it also demonstrates $\{P_n(X)\}_{n=1}^{N} \subseteq (\{P_n\}_{n=1}^{\infty}, +, \times)$ e.g., $P_n(X) \stackrel{\text{def}}{=} \prod_{i=1}^{N} x_i + \sum_{j=1}^{N} y_j$ [36, 37].

If $\mathfrak{F}$ is a real function denoted on set $\mathcal{F}$, it indicates:

$$\lim_{N\to\infty}\left|\mathfrak{F} - \{P_n(X)\}_{n=1}^{N}\right| = 0 \tag{A.12}$$

then we have $\lim_{N\to\infty}\{P_n(X)\}_{n=1}^{N} = \mathfrak{F}$, meanwhile, if N is large enough, $\left|\mathfrak{F} - \{P_n(X)\}_{n=1}^{N}\right| < \varepsilon$ holds.

***Lemma 1.2* (Contraction of Operators Combination)** Given two contraction mappings $\Phi_1$ and $\Phi_2$, we have the composite of two contraction mapping as $\Phi_2 \cdot \Phi_1$. The composite mapping $\Phi_2 \cdot \Phi_1$ must be contractive.

***Proof***: According to the definition of contraction linear operator, we have:

$$\exists \zeta \in (0,1)$$
$$\rho \stackrel{\text{def}}{=} \|\Phi x - \Phi y\| \tag{A.13}$$
$$\rho(\Phi x, \Phi y) \le \zeta \rho(x, y)$$

Obviously, and we have:

$$\rho(\Phi_1 u, \Phi_1 v) \le \zeta \rho(u, v) \ \forall \zeta \in (0,1) \tag{A.14}$$
$$\rho(\Phi_2 x, \Phi_2 y) \le \eta \rho(x, y) \ \forall \eta \in (0,1)$$

If we set:

$$x = \Phi_1 u, y = \Phi_1 v \tag{A.15}$$

the inequality below holds:

$$\rho(\Phi_2 x, \Phi_2 y) \le \eta \rho(\Phi_1 u, \Phi_1 v) \le \zeta \eta \rho(u, v) \tag{A.16}$$

Since the definition as

$$\forall \zeta, \eta \in (0,1), \rho(\Phi_2 \Phi_1 u, \Phi_2 \Phi_1 y) \le \zeta \eta \rho(u, v) \tag{A.17}$$

***Corollary 1.1* (General Contraction Operator)** According to Lemma 1.2, if denote the operators $\{\Phi_i\}_{i=1}^{K}$, $\forall \Phi_i \ i \in \mathbb{N}, \Phi_i: \mathbb{R}^{S\times T} \to \mathbb{R}^{S\times T}$; considering any combination of operators: $\Phi_K \cdot \cdots \cdot \Phi_2 \cdot \Phi_1$, if at least a single operator $\Phi_i$ is contraction operator, and other operators are bounded, such as $\forall i \ne k \ \|\Phi_i\| \le M$. If and only if $\prod_{i=1}^{K}\|\Phi_i\| < 1$, the combination of operator series $\Phi_K \cdot \cdots \cdot \Phi_2 \cdot \Phi_1$ is a contraction operator.

***Proof***: Obviously, according to Lemma 1.2, use a series as $\{\zeta_i\}_{i=1}^{K}$ to replace $\zeta, \eta \in (0,1)$,

Obviously, we have:

$$\zeta_i \in (0,1) \; i \in \mathbb{N} \tag{A.18}$$

$$\rho(\Phi_K \cdot \cdots \cdot \Phi_2 \Phi_1 u, \Phi_K \cdot \cdots \cdot \Phi_2 \Phi_1 y) \leq \zeta_K \cdot \cdots \zeta_2 \cdot \zeta_1 \cdot \rho(u,v)$$

Since $\zeta_K \cdot \cdots \zeta_2 \cdot \zeta_1 < 1$, we have proved this corollary.

***Corollary 1.2* (Iterative Contraction Operator)** According to Lemma 1.2, if denote the operators $\{\Phi_i\}_{i=1}^K$, $\forall \Phi_i \; i \in \mathbb{N}$, $\Phi_i: \mathbb{R}^{S \times T} \to \mathbb{R}^{S \times T}$; considering any combination of operators: $\Phi_K \cdot \cdots \cdot \Phi_2 \cdot \Phi_1$, if at least a single operator $\Phi_i$ is contraction operator, and other operators are bounded, such as $\forall i \neq k, \|\Phi_i\| \leq M$. If and only if $\lim_{n \to \infty} \prod_{i=1}^K \|\Phi_i\|^n = c < 1$, the combination of operator series $\Phi_K^n \cdot \cdots \cdot \Phi_2^n \cdot \Phi_1^n$.

***Proof***: Obviously, according to Lemma 1.2 and Corollary 1.1 and 1.2, use a series as $\{\zeta_i\}_{i=1}^K$ to replace $\zeta, \eta \in (0,1)$,

And we have:

$$\forall \zeta_i \in (0,1) \; i \in \mathbb{N} \tag{A.19}$$

$$\rho(\Phi_K^n \cdot \cdots \cdot \Phi_2^n \cdot \Phi_1^n u, \Phi_K^n \cdot \cdots \cdot \Phi_2^n \cdot \Phi_1^n y) < \zeta_i^n \cdot \cdots \cdot \zeta_2^n \cdot \zeta_1^n \cdot \rho(u,v)$$

Since $0 < \zeta_i^n \cdot \cdots \cdot \zeta_2^n \cdot \zeta_1^n < 1$, we have proved this corollary.

***Lemma 1.3* (Convex of Norm)** Any norm of finite dimensional matrix is convex.

***Proof***: For any finite dimensional real matrix $X \subseteq R^{M \times N}$, according to inequality of norm, we have:

$$\|X\| = \|(1 - \rho + \rho)X\| \leq \rho\|X\| + (1-\rho)\|X\|$$

We have proved that $\|X\|$ is convex.

***Lemma 1.4* (Convex of Alternative Optimization of Proposed Function)**

***Proof***: To be simplified, we prove all involved variables in a single layer and assume all variables are completely independent.

$S$ is input signal matrix and a constant; at first, we assume $Z$ and $Y$ as two constant variables, and $\forall \varepsilon > 0$ $\|Z - S\| < \varepsilon$, then we have:

$$\|(1 - \rho + \rho)XY + Z - S\| \leq \rho\|X\| \cdot \|Y\| + (1-\rho)\|X\| \cdot \|Y\| + \|Z - S\| \tag{A.20}$$

Due to the randomness of $\varepsilon$, the following equation holds:

$$\|(1 - \rho + \rho)XY\| \leq \rho\|X\| \cdot \|Y\| + (1-\rho)\|X\| \cdot \|Y\| \tag{A.21}$$

Since $Y$ is already treated as a constant, according to Lemma 1.3, we have proved the alternative function for $X$ is convex.

Similarly, according to Lemma 1.3, we can prove the alternative strategy for $Y$ is convex.

Then, we assume $\forall \varepsilon > 0 \; \|XY - S\| < \varepsilon$ and we have:

$$\|(1 - \rho + \rho)Z + XY - S\| \leq \rho\|Z\| + (1 - \rho)\|Z\| + \|XY - S\| \tag{A.22}$$

Due to $\forall \varepsilon > 0\ \|Z - S\| < \varepsilon$, then we have:

$$\|(1 - \rho + \rho)Z\| \leq \rho\|Z\| + (1 - \rho)\|Z\| \tag{A.23}$$

Therefore, we have proved the alternative optimization function of $Z$ is also convex.

***Lemma 1.5* (Convergence of Alternative Optimization of Real Function)** For a series of real function as $\{f_{i,j}\}_{i,j=1}^{\infty}$. If we have: $\lim_{i \to \infty} f_{i,j}(x) \to h_{M,j}, a.e.\ x \in [a,b]$ and $\lim_{j \to \infty} h_{M,j} \to g_{M,N}, a.e.\ x \in [a,b]$. Then, $\exists \lim_{k \to \infty} f_{i_k,j_k} \to g_{M,N}\ a.e.\ x \in [a,b]$ holds.

***Proof***: If considering the uniform convergence of $\{f_{i,j}\}_{i,j=1}^{\infty}$, since $\lim_{i \to \infty} f_{i,j} \to h_{M,j}, a.e.\ x \in [0,1]$ and $\lim_{j \to \infty} h_{M,j} \to g_{M,N}, a.e.\ x \in [0,1]$, according to Riez Theorem, $\exists \{f_{i_k,j_k}\}_{k=1}^{\infty}$.

$$\left|f_{i_k,j_k} - h_j\right| < \frac{\varepsilon}{2}, \left|h_j - g\right| < \frac{\varepsilon}{2} \tag{A.23}$$

Then, we have:

$$\left|f_{i_k,j_k} - h_j\right| + \left|h_j - g\right| = \left|f_{i_k,j_k} - g\right| < \varepsilon \tag{A.24}$$

***Theorem 1.2* (Alternative Convergence of Deep Linear Models)** If considering the continuous operator applied on finite dimensional space, the series of operators, $\{\mathcal{F}_{i,j}\}_{i,j=1}^{\infty}$, $\{\mathcal{H}_{K,j}\}_{j=1}^{\infty}$. And $\mathcal{F}_{i,j}: \mathbb{R}^{M \times N} \to \mathbb{R}^{M \times N}$. $\mathcal{H}_{K,j}: \mathbb{R}^{M \times N} \to \mathbb{R}^{M \times N}$. If we have: $\lim_{i \to \infty} \mathcal{F}_{i,j} \to \mathcal{H}_{K,j}$ and $\lim_{j \to \infty} \mathcal{H}_{K,j} \to \mathcal{G}$. Then, $\exists \lim_{k \to \infty} \mathcal{F}_{i_k,j_k} \to \mathcal{G}$ holds.

***Proof***: According to Lemma 1.5, similarly, let constant $T < \infty$, we have:

$$\begin{aligned}\left\|\mathcal{F}_{i_k,j_k} - \mathcal{H}_{K,j}\right\| &< \frac{\varepsilon}{2T}\\ \left\|\mathcal{H}_{K,j} - \mathcal{G}\right\| &< \frac{\varepsilon}{2T}\end{aligned} \tag{A.26}$$

The following inequality holds:

$$\left\|\mathcal{F}_{i_k,j_k} - \mathcal{H}_{K,j}\right\| + \left\|\mathcal{H}_{K,j} - \mathcal{G}\right\| = \left\|\mathcal{F}_{i_k,j_k} - \mathcal{G}\right\| < \frac{\varepsilon}{T} \tag{A.27}$$

And we also have:

$$\left\|\mathcal{F}_{i_k,j_k} X - \mathcal{G}X\right\| \leq \left\|\mathcal{F}_{i_k,j_k} - \mathcal{G}\right\| \cdot \|X\| < T \cdot \frac{\varepsilon}{T} = \varepsilon \tag{A.28}$$

In details, if operator $\mathcal{F}_{i_k,j_k}$ and $\mathcal{H}_{K,j}$ denote the optimization operator on variables $X$ and $Y$ in Eq. (1), obviously, via alternative optimization strategy, the result indicates the operator can converge to a fixed point defined on Banach space, using alternative strategy and Banach Fixed Point Theorem, if and only if $\lim_{k \to \infty} \left\|\mathcal{F}_{i_k,j_k}\right\| < 1$.

*Strategy of Orthogonal Acceleration*

We rewrite the Augmented Lagrangian Function as a single layer $k$:

$$\mathcal{L}_\beta(X_k, Y_k, Z_k, e_k) \stackrel{\text{def}}{=} \frac{\beta}{2}\|X_k Y_k - S\|_F^2 + \langle X_k Y_k - S, e_k\rangle + \frac{1}{\beta}\|Z_k\|_1 \tag{A.29}$$

Obviously, via calculating the partial derivatives of $X_k$ and $Y_k$, in Eqs. (A.30) and (A.31), respectively:

$$\frac{\partial \mathcal{L}_\beta}{\partial Y_k} = \beta(X_k Y_k - S)X_k^T + e_k X_k^T \tag{A.30}$$

For partial derivative of $X_k$:

$$\frac{\partial \mathcal{L}_\beta}{\partial X_k} = \beta(X_k Y_k - S)Y_k^T + e_k Y_k^T \tag{A.31}$$

Let these two derivatives of $X_k$ and $Y_k$ be 0, then we have:

$$\frac{\partial \mathcal{L}_\beta}{\partial Y_k} = 0 \Rightarrow Y_k = X_k^\dagger (S - \frac{e_k}{\beta}) \tag{A.32}$$

Similarly, we also have:

$$\frac{\partial \mathcal{L}_\beta}{\partial X_k} = 0 \Rightarrow X_k = (S - \frac{e_k}{\beta})Y_k^\dagger \tag{A.33}$$

In [], using Eq. (A.32), and orthogonal projection technique introduced in [], we can derive:

$$X_k Y_k = X_k X_k^\dagger \left(S - \frac{e_k}{\beta}\right) = \mathcal{P}_{X_k}\left(S - \frac{e_k}{\beta}\right) \tag{A.34}$$

Furthermore, obviously, due to $rank(X_k)$ is equal to $rank((S - \frac{e_k}{\beta})Y_k^T)$, and denoting $Q$ as the orthogonal basis of $(S - \frac{e_k}{\beta})Y_k^T$, we can conclude:

$$X_k Y_k = \mathcal{P}_{(S - \frac{e_k}{\beta})Y_k^T}\left(S - \frac{e_k}{\beta}\right) = QQ^T\left(S - \frac{e_k}{\beta}\right) \tag{A.35}$$

Thus, the accelerated iterative format can be written as Eqs. (A.36) and (A.37):

$$X_{k+1} \leftarrow QR\left(\left(S - \frac{e_k}{\beta}\right)Y_k^T\right) \tag{A.36}$$

$$Y_{k+1} \leftarrow QR\left(X_k^T\left(S - \frac{e_k}{\beta}\right)\right) \tag{A.37}$$

# Appendix B

***Definition 2.1*** If we denote DELMAR as an operator $\mathfrak{U}$, based on the description of DELMAR, considering the iteration k, we can denote $\mathfrak{U} \stackrel{\text{def}}{=} M \cdot \mathcal{A}^k \cdot \mathcal{S}^k \cdot \mathcal{R}^k$.

***Definition 2.2*** If we denote Deep SDL as an operator $\mathfrak{L}$, based on the description of Deep SDL, considering the iteration k, we can denote $\mathfrak{L} \stackrel{\text{def}}{=} M \cdot \mathcal{G}^k \cdot \mathcal{S}^k$.

***Definition 2.3*** If we denote Deep FICA as an operator $\mathfrak{T}$, based on the description of Deep FICA, considering the iteration k, we can denote $\mathfrak{T} \stackrel{\text{def}}{=} \mathcal{P} \cdot \mathcal{F}^k$.

***Definition 2.4*** If we denote Deep NMF as an operator $\mathfrak{N}$, based on the description of Deep NMF, considering the iteration k, we can denote $\mathfrak{N} \stackrel{\text{def}}{=} M \cdot \mathcal{U}^k \cdot \mathcal{N}$.

***Theorem 2.1*** **(Contraction of ADMM Operator)** ADMM could be considered as contraction operator. It can be treated as a general iterative contraction operator in finite dimensionality space. We have $ADMM \stackrel{\text{def}}{=} \mathcal{A}$. If denote the $\|\mathcal{A}^{k+1}\| = \alpha\|\mathcal{A}^k\|$, and $\beta$ should be step length, i.e., penalty parameter, if $n \to \infty$ $0 < (\alpha\beta)^n \|BN\| < 1$, $\mathcal{A}$ can be considered as a contraction operator. And $\|BN\|$ denotes the norm of different residual error, considering two distinctive input matrices.

***Proof***: $X$ and $Y$, represent the two input matrices.

Consider the iterative format of ADMM as

$$\mathcal{A}_{k+1} \leftarrow \mathcal{A}_k - \min(f_\mathcal{A}) \tag{B.1}$$

And it also can imply:

$$\|\mathcal{A}_{k+1}\| = \alpha\|\mathcal{A}_k\|, \tag{B.2}$$
$$0 < \alpha < 1$$

According to the definition of contraction operator, we have:

$$\|\mathcal{A}X - \mathcal{A}Y\| \leq \alpha \left\| \left( \beta(e_k^t + \prod_{i=1}^{k-1} X_i Y_k + \sum_{i=1}^{k} Z_k^{t+1} - S) - \alpha\beta(\hat{e}_k^t + \prod_{i=1}^{k-1} \hat{X}_i \hat{Y}_k + \sum_{i=1}^{k} \hat{Z}_k^{t+1} - \hat{S}) \right) \right\| \tag{B.3}$$

And we also have:

$$\|\mathcal{A}X - \mathcal{A}Y\| \leq \alpha\beta \left\| e_k^t - \hat{e}_k^t + \prod_{i=1}^{k-1} X_i Y_k - \prod_{i=1}^{k-1} \hat{X}_i \hat{Y}_k + \sum_{i=1}^{k} Z_k^{t+1} - \sum_{i=1}^{k} \hat{Z}_k^{t+1} + \hat{S} - S \right\| \tag{B.4}$$

Since $e_k^t, \hat{e}_k^t, \prod_{i=1}^{k-1} X_i Y_k, \prod_{i=1}^{k-1} \hat{X}_i \hat{Y}_k, \sum_{i=1}^{k} Z_k^{t+1}, \sum_{i=1}^{k} \hat{Z}_k^{t+1}, \hat{S}, S \in \mathbb{R}^{m \times n}$, they are obviously bounded; and using Corollary 1.1 and 1.2, we have:

$$\|\mathcal{A}X - \mathcal{A}Y\| \leq \alpha\beta \|BN\| \tag{B.5}$$

Obviously, it demonstrates:

$$\|\mathcal{A}^n A - \mathcal{A}^n B\| \leq (\alpha\beta)^n \|BN\| < 1 \tag{B.6}$$

If and only if $0 < (\alpha\beta)^n < 1$, or $0 < \alpha\beta < 1$, $\mathfrak{A}^n$ is equivalent to a contraction operator. According to Lemma 1.2 and Corollary 1.1, 1.2, it also indicates: when $n$ is large enough, $n > N$, we have:

$$\lim_{n\to\infty} \|\mathcal{A}^n A - \mathcal{A}^n B\| \leq \lim_{n\to\infty} (\alpha\beta)^n \|BN\| \tag{B.7}$$

Obviously, if and only if $\lim_{n\to\infty} (\alpha\beta)^n \|BN\| < 1$, the iterative ADMM operator can be equivalent to a contraction operator.

***Theorem 2.2* (Initialization Operator is bounded)** If we denote the sparse operator as $\mathcal{M}: \mathbb{R}^{S \times T} \to \mathbb{R}^{S \times T}$, we have $\|\mathcal{M}\| < \infty$.

***Proof***: according to the definition of operator norm [30], $\|\mathcal{M}\| \leq \sup \frac{\|\mathcal{M}X\|}{\|X\|}$; obviously, $\|\mathcal{M}X\|$ and $\|X\|$ is bounded, since both of norms are based on finite dimensional matrix. And if we denote:

$$X = \begin{bmatrix} a_1 \\ a_2 \\ \vdots \\ a_{n-1} \\ a_n \end{bmatrix} \quad \mathcal{M}X = \begin{bmatrix} b_1 \\ b_2 \\ \vdots \\ b_{n-1} \\ b_n \end{bmatrix} \quad \|X\| < \infty \quad \|\mathcal{M}X\| < \infty \tag{B.8}$$

Obviously, $\|\mathcal{M}\| < \infty$.

***Theorem 2.3* (Sparsity Operator is bounded)** If we denote the sparse operator as $\mathcal{S}: \mathbb{R}^{S \times T} \to \mathbb{R}^{S \times T}$, we have $\|\mathcal{S}\| < \infty$.

***Proof***: according to the definition of operator norm [30], $\|\mathcal{S}\| \leq \sup \frac{\|\mathcal{S}X\|}{\|X\|}$; obviously, $\|\mathcal{S}X\|$ and $\|X\|$ is bounded, since both of norms are based on finite dimensional matrix. And if we denote:

$$\mathcal{S}X = \begin{bmatrix} a_1 \\ a_2 \\ \vdots \\ 0 \\ a_n \end{bmatrix} \quad \mathcal{S}Y = \begin{bmatrix} b_1 \\ 0 \\ \vdots \\ b_{n-1} \\ b_n \end{bmatrix} \tag{B.9}$$

and we examine:

$$SX - SY = \begin{bmatrix} a_1 - b_1 \\ a_2 \\ \vdots \\ -b_{n-1} \\ a_n - b_n \end{bmatrix}; X - Y = \begin{bmatrix} a_1 - b_1 \\ a_2 - b_2 \\ \vdots \\ a_{n-1} - b_{n-1} \\ a_n - b_n \end{bmatrix}, \|SX - SY\| \leq s \|X - Y\|, \tag{B.10}$$

Without loss of generality, and based on Lemma 1.2, we calculate the $\ell_2$ norm, and we have:

$$\infty > s \geq \frac{\sum_{i=u}^{n}(a_i - b_i)^2 + \sum_{i=v}^{p}(a_i)^2 + \sum_{i=w}^{t}(b_i)^2}{\sum_{i=1}^{n}(a_i - b_i)^2} \tag{B.11}$$

This inequality demonstrates that $\|S\|$ is a bounded. And $S$ is a bounded operator.

***Theorem 2.4*** **(Rank Reduction Operator is bounded)** If we denote the sparse operator as $\mathcal{R}: \mathbb{R}^{S \times T} \to \mathbb{R}^{S \times T}$, we have $\|\mathcal{R}\| < \infty$.

***Proof***: According to the definition of operator norm [30], $\|\mathcal{R}\| \leq sup \frac{\|\mathcal{R}X\|}{\|X\|}$; obviously, $\|\mathcal{R}X\|$ and $\|X\|$ is bounded, since both of norms are based on finite dimensional matrix. And if we denote:

$$X = \begin{bmatrix} a_1 \\ a_2 \\ \vdots \\ a_{n-1} \\ a_n \end{bmatrix}, \mathcal{R}X = \begin{bmatrix} b_1 \\ b_2 \\ \vdots \\ b_k \\ \vdots \\ 0 \end{bmatrix} \tag{B.12}$$

Eq. (B12) implies:

$$sup \frac{\|\mathcal{R}X\|}{\|X\|} = \frac{\sum_{i=1}^{n} a_i^2}{\sum_{i=u}^{p}(a_i - b_i)^2 + \sum_{i=v}^{q} a_i^2} < \infty. \tag{B.13}$$

Also, if we examine the weighted ratio and weight difference, only considering the finite dimensional space, we have:

$$X = \begin{bmatrix} a_1 \\ a_2 \\ \vdots \\ a_{n-1} \\ a_n \end{bmatrix}, WR \cdot X = \begin{bmatrix} a_2/a_1 \\ a_3/a_2 \\ \vdots \\ a_k/a_{k-1} \\ \vdots \\ a_n/a_{n-1} \; 0 \end{bmatrix} WD \cdot X = \begin{bmatrix} a_2 - a_1 \\ a_3 - a_2 \\ \vdots \\ a_k - a_{k-1} \\ \vdots \\ a_n - a_{n-1} \end{bmatrix}$$

Obviously, for each rank estimation, the dimension of input matrix can be reduced at least by one. Similarly, WR and WD can be considered as the contract operators for dimensional estimation. It demonstrate that the input matrix can be reduced to a vector by n-1 iterations at most.

***Theorem 2.5*** **(Normalization Operator of Deep NMF is bounded)** If we denote the normalization operator of Deep NMF as $\mathcal{N}: \mathbb{R}^{S \times T} \to \mathbb{R}^{S \times T}$, we have $\|\mathcal{N}\| \leq 1$.

***Proof***: according to the definition of operator norm [30], $\|\mathcal{N}\| \leq sup \frac{\|\mathcal{N}X\|}{\|X\|}$; obviously, $\|\mathcal{N}X\|$ and $\|X\|$ is bounded, since both of norms are based on finite dimensional matrix. And if we denote:

$$X = \begin{bmatrix} a_1 \\ a_2 \\ \vdots \\ a_{n-1} \\ a_n \end{bmatrix} \mathcal{N}X = \begin{bmatrix} b_1 \\ 0 \\ \vdots \\ b_{n-1} \\ b_n \end{bmatrix} \quad (B.14)$$

According to Eq. (B.14), we need to notice: $\{a_i\}_{i=1}^K \subseteq [-q, q]$, $1 \leq q < \infty$; $\{b_i\}_{i=1}^K \subseteq [0,1]$. Obviously, $\|\mathcal{N}X\| < \|X\|$. Finally, we have: $\|\mathcal{N}\| < 1$.

***Theorem 2.6 (Contraction of Updating Operator Deep NMF)*** If we denote the updating operator as $\mathcal{U}: \mathbb{R}^{S \times T} \to \mathbb{R}^{S \times T}$, we have $\|\mathcal{U}\| < 1$.

***Proof***: according to the definition of operator norm [30], $\|\mathcal{U}\| \leq sup \frac{\|\mathcal{U}X\|}{\|X\|}$; obviously, $\|\mathcal{U}X\|$ and $\|X\|$ is bounded, since both of norms are based on finite dimensional matrix. And if we denote:

$$X = \begin{bmatrix} a_1 \\ a_2 \\ \vdots \\ a_{n-1} \\ a_n \end{bmatrix} \mathcal{U}X = \begin{bmatrix} b_1 \\ 0 \\ \vdots \\ b_{n-1} \\ b_n \end{bmatrix} \quad (B.15)$$

According to the iterative format of Deep NMF, we need to notice: $b_i = \frac{a_i}{\max f(a_i)}$; obviously, $\|\mathcal{U}X\| < \|X\|$. Finally, we have: $\|\mathcal{U}\| < 1$. Otherwise, if $\|\mathcal{U}\| > 1$, when $k \to \infty$, we have: $\|\mathcal{U}X\| = \infty$.

***Theorem 2.7 (Contraction of GD Operator)*** Gradient Descent (GD) is a bounded contraction operator, if and only if the derivative of target function is bounded:

$$|f''(\varsigma)| < \frac{1}{\sigma} < \infty, \sigma \text{ is the step length.}$$

***Proof***: The standard iteration format is:

$$x_{k+1} = x_k - \sigma f'(x_k) \quad (B.16)$$

Using the definition of operator, we have:

$$\tau(x_k) = x_k - \sigma f'(x_k) \, \forall \sigma \in (0,1) \quad (B.17)$$

And we have:

$$\|\tau X - \tau Y\| = \|(X - Y) - \sigma(f'(X) - f'(Y))\| \quad (B.18)$$

Using Mean value theorem, we have:

$$\|\tau X - \tau Y\| = |1 - \sigma f''(\varsigma)|\|X - Y\| \tag{B.19}$$

According to the definition of contraction operator [30], if and only if:

$$|1 - \sigma f''(\varsigma)| < 1, |1 - \sigma f''(\varsigma)| \in \mathbb{K} \tag{B.20}$$

It also implies, when the following inequality holds:

$$|f''(\varsigma)| < \frac{1}{\sigma} < \infty \tag{B.21}$$

GD is considered as a contraction mapping/operator. Without generality, we can set $\sigma < \frac{1}{|f''(x)|+1}$.

And obviously, using multiplicative inequality, we have:

$$\|\tau X - \tau Y\| \leq \|\tau\|\|X - Y\| \tag{B.22}$$

Since $X$ and $Y$ both denote in finite $\ell^2$ space, we have:

$$\|\tau\|\|X - Y\| \leq \infty \tag{B.23}$$

Using Uniformly bounded theorem, we have:

$$\|\tau\| \leq M, M \in \mathbb{K} \tag{B.24}$$

GD is a bounded mapping/operator.

According to Lemma 1.2, and Corollary 1.1-1.2, obviously, for $n$ iterations for an operator, and if we set the accuracy level as $\varepsilon$, we have:

$$\|\tau^n X - \tau^{n+1} Y\| = \sigma^n \|X - \tau Y\| < \varepsilon \tag{B.25}$$

Since $X$ and $Y$ is both denoted in finite $\ell^2$ space, we have:

$$\sigma^n \|X - \tau Y\| \leq \sigma^n (\|X\| + \|\tau Y\|) \tag{B.26}$$

Obviously, $\|X - Y\|_{\ell^2}$ is bounded, and we have:

$$\sigma^n(\|X\| + \|\tau Y\|) \leq \sigma^n(\|X\| + \|\tau\|\|Y\|) \leq \sigma^n(\|X\| + \|Y\|) \leq \sigma^n \cdot 2\|X\|$$

$$0 < \sigma^n \cdot 2\|X\| < \varepsilon \tag{B.27}$$

$$n > \log \frac{\varepsilon}{2\|X\|} / \log \sigma > 0$$

We provide the infimum of iteration as $\log \frac{\varepsilon}{2\|X\|} / \log \sigma$ to approach the accuracy level $\varepsilon$.

***Theorem 2.8 (Operator PCA is bounded)*** If we denote the updating operator as $\mathcal{P}: \mathbb{R}^{S \times T} \to \mathbb{R}^{S \times T}$, we have $\|\mathcal{P}\| < \infty$.

***Proof***: According to the definition of operator norm [30], $\|\mathcal{P}\| \leq \sup \frac{\|\mathcal{P}X\|}{\|X\|}$; obviously, $\|\mathcal{U}X\|$ and $\|X\|$ is bounded, since both of norms are based on finite dimensional matrix. And if we denote:

$$X = \begin{bmatrix} a_1 \\ a_2 \\ \vdots \\ a_{n-1} \\ a_n \end{bmatrix} \mathcal{P}X = \begin{bmatrix} b_1 \\ b_2 \\ \vdots \\ b_{n-k} \\ \vdots \\ 0 \end{bmatrix} \quad (B.29)$$

According to the dimensional reduction of PCA, we have: $sup \frac{\|\mathcal{P}X\|}{\|X\|} = sup \frac{(\sum_{i=1}^{n-k} b_i^2)^{\frac{1}{2}}}{(\sum_{i=1}^{n} a_i^2)^{\frac{1}{2}}} < \infty$. It demonstrates: $\|\mathcal{P}\| < \infty$.

***Theorem 2.9*** **(Contraction of Fixed-Point Operator)** If we denote the updating operator as $\mathcal{F}: \mathbb{R}^{S \times T} \to \mathbb{R}^{S \times T}$, we have $\|\mathcal{F}\| < 1$.

***Proof***: according to the definition of operator norm [30], $\|\mathcal{F}\| \leq sup \frac{\|\mathcal{F}X\|}{\|X\|}$; obviously, $\|\mathcal{U}X\|$ and $\|X\|$ is bounded, since both of norms are based on finite dimensional matrix. And if we denote:

$$X = \begin{bmatrix} a_1 \\ a_2 \\ \vdots \\ a_{n-1} \\ a_n \end{bmatrix} \mathcal{F}X = \begin{bmatrix} b_1 \\ 0 \\ \vdots \\ b_{n-1} \\ b_n \end{bmatrix} \quad (B.30)$$

According to the iterative format of Deep NMF, we need to notice: $b_i = \frac{a_i}{\sqrt{\|a_i\| C_i \|a_i^T\|}}$; obviously, $\|\mathcal{F}X\| < \|X\|$. Finally, we have: $\|\mathcal{F}\| < 1$. Otherwise, if $\|\mathcal{F}\| > 1$, when $k \to \infty$, we have: $\|\mathcal{F}X\| = \infty$.

***Theorem 2.10*** **(Inequality of Operator Norms)** According to Theorem 2.1, 2.7, 2.8 and 3.0, if we assume: $\|\mathcal{A}^{k+1}\| = \alpha_1 \|\mathcal{A}^k\|, \|\mathcal{G}^{k+1}\| = \alpha_2 \|\mathcal{G}^k\|, \|\mathcal{U}^{k+1}\| = \alpha_3 \|\mathcal{U}^k\|, \|\mathcal{F}^{k+1}\| = \alpha_4 \|\mathcal{F}^k\|$, we have: $\alpha_1 \neq \alpha_3, \alpha_4$; $\alpha_2 \neq \alpha_3, \alpha_4$;

***Proof***: Proof by contradiction. In general, we assume $\|\mathcal{A}\| = \|\mathcal{U}\|$, according to the iterative formats of DELMAR and Deep NMF, and considering:

$$\mathcal{A}_{k+1} \leftarrow \mathcal{A}_k - \min(f_\mathcal{A}) \quad (B.31)$$

If we employ the $\alpha \mathcal{A}_k = \mathcal{A}_{k+1}$ to replace $\mathcal{A}_{k+1}$:

$$ \quad (B.32)$$

And we can reformat this equality as:

$$(1 - \alpha)\mathcal{A}_k = \min(f_\mathcal{A}) \quad (B.33)$$

Considering the iterative format of Deep NMF:

$$\mathfrak{N}_{k+1} \leftarrow \mathfrak{N}_k / \max(f_\mathfrak{N}) \quad (B.34)$$

Let we denote:

$$\left|\frac{1}{max(f_\mathfrak{N})}\right| \leq \varepsilon \tag{B.35}$$

And considering an extreme condition, $\forall \varepsilon_i \leq \varepsilon$, for each iteration $i$, and $\lim_{i\to\infty} \varepsilon_i = \varepsilon$;

$$\lim_{i\to\infty}(1-\varepsilon_i)\mathcal{A}_k = \min(f_\mathcal{A}) \tag{B.36}$$

Then we have the conclusion:

$$\exists\, n \ll k, \mathcal{A}_k = \min(f_\mathcal{A}) \tag{B.37}$$

It demonstrates for the iterative format of DELMAR, before convergence, the iteration can be terminated, since a very small norm of operator $\mathcal{A}$. $\mathcal{A}$ cannot guarantee the convergence. It obviously disobeys the property of ADMM.

Similarly, we can also prove $\alpha_2 \neq \alpha_3, \alpha_4$; and $\alpha_1 \neq \alpha_4$.

# Appendix C

***Assumption 3.1*** For all operators, these operators should be considered as linear operators, and we have:

$$\Phi \cdot (X + Y) = \Phi \cdot X + \Phi \cdot Y \tag{C.1}$$

Otherwise**,**

$$\Phi \cdot (X + Y) \neq \Phi \cdot X + \Phi \cdot Y \tag{C.2}$$

Most operators can be considered as the linear operator and satisfy Eq. (C.1) [30].

***Assumption 3.2*** For any input matrix, we can successfully separate the vital information and background noise. If we denote: $V = \{\cup_{i=1}^{P} voxel_i, voxel_i \in BN\}$, and $N = \{\cup_{i=1}^{Q} voxel_i, voxel_i \notin BN\}$. BN represents the functional areas, i.e., potentially activated areas of brain. We have some crucial assumptions: $V \cap N = \emptyset, V \geqslant 0, B \geqslant 0, \|V\| \gg \|N\|$.

***Lemma 3.1*** **(Continuous Operators)** For all operators analyzed in this study, if $k > K, \forall k \in \mathbb{N}$, these iterative operators can be considered as consistent operator. It means: if we have $\|V - \hat{V}\| \leq \varepsilon$, $\|\mathfrak{A}^k V - \mathfrak{A}^k \hat{V}\| \to 0$.

***Proof***: We denote: $\mathfrak{A}, \mathfrak{L}, \mathfrak{T}, \mathfrak{N} \in \mathfrak{C}: \mathbb{R}^{s \times t} \to \mathbb{R}^{s \times t}$

For $V, \hat{V} \subseteq \mathbb{R}^{s \times t}$, we assume that:

$$\|V - \hat{V}\| \leq \frac{\varepsilon}{M} \tag{C.3}$$

If $k > K$, For any operator belongs to $\mathfrak{C}$ can be considered as a contraction operator, and we have:

$$\|\mathfrak{A}^k V - \mathfrak{A}^k \hat{V}\| \leq \|\mathfrak{A}^k\| \cdot \|V - \hat{V}\| \leq M \cdot \frac{\varepsilon}{M} = \varepsilon \tag{C.4}$$

This inequality demonstrates that all operators of $\mathfrak{C}$, if $k$ is large enough, can be treated as the consistent operators . Similarly, it also demonstrates: $\|\mathfrak{A}^k V - \mathfrak{L}^k V\| \leq \varepsilon$

***Theorem 3.1*** **(Distinctive Spatial Similarity)** If we denote the following set:

$$\begin{aligned} Deep\ MF: D &= \{\mathfrak{A}^k N, N \in \bigcup_{i=1}^{M} voxel_i, voxel_i \notin T\} \\ Deep\ SDL: L &= \{\mathfrak{L}^k N, N \in \bigcup_{i=1}^{M} voxel_i, voxel_i \notin T\} \\ Deep\ FICA: I &= \{\mathfrak{T}^k N, N \in \bigcup_{i=1}^{M} voxel_i, voxel_i \notin T\} \end{aligned} \tag{C.5}$$

$$Deep\ NMF: \Theta = \{\mathfrak{N}^k N, N \in \bigcup_{i=1}^{M} voxel_i,\ voxel_i \to 0\}$$

And considering the iteration *k*, it implies:

$$\frac{|\mathfrak{A}^k V|}{|V \cup D|} \leq \frac{|\mathfrak{L}^k V|}{|V \cup L|} \leq \frac{|\mathfrak{T}^k V|}{|V \cup I|} \leq \frac{|\mathfrak{N}^k V|}{|V \cup \Theta|} \tag{C.6}$$

where $|\cdot|$ denotes the number of positive elements.

***Proof***: Based on assumptions 3.1 and 3.2, if $\forall k \in \mathbb{N}$, we have:

$$\begin{aligned}
\mathfrak{A}^k C &= \mathfrak{A}^k V + (\mathfrak{A}^k N) \\
\mathfrak{L}^k C &= \mathfrak{L}^k V + (\mathfrak{L}^k N) \\
\mathfrak{T}^k C &= \mathfrak{T}^k V + (\mathfrak{T}^k N) \\
\mathfrak{N}^k C &= \mathfrak{N}^k V + \Theta
\end{aligned} \tag{C.7}$$

According to Corollary 1.1 and 1.2, $k > K$, we have:

$$0 = \|\Theta\| \leq \|\mathfrak{T}^k N\| \leq \|\mathfrak{L}^k N\| \leq \|\mathfrak{A}^k N\| < \infty \tag{C.8}$$

We can also rewrite it as:

$$0 = |\Theta| < |I| \leq |L| \leq |D| \tag{C.9}$$

And, according to the spatial similarity, we also have:

$$\begin{aligned}
Deep\ MF_{Similarity} &\stackrel{def}{=} \frac{|(\mathfrak{A}^k V \cup D) \cap V|}{|V \cup (\mathfrak{A}^k A \cup D)|} = \frac{|\mathfrak{A}^k V|}{|V \cup D|} \\
Deep\ SDL_{Similarity} &\stackrel{def}{=} \frac{|(\mathfrak{L}^k V \cup L) \cap A|}{|V \cup (\mathfrak{L}^k V \cup L)|} = \frac{|\mathfrak{L}^k A|}{|V \cup L|} \\
Deep\ FICA_{Similarity} &\stackrel{def}{=} \frac{|(\mathfrak{T}^k V \cup I) \cap V|}{|V \cup (\mathfrak{T}^k A \cup I)|} = \frac{|\mathfrak{T}^k V|}{|V \cup I|} \\
Deep\ NMF_{Similarity} &\stackrel{def}{=} \frac{|(\mathfrak{N}^k V \cup \Theta) \cap V|}{|V \cup (\mathfrak{N}^k A \cup \Theta)|} = \frac{|\mathfrak{N}^k V|}{|V|}
\end{aligned} \tag{C.10}$$

Again, considering $k > K$, and Corollary 1.1 to 1.2, and Theorem 3.2, we have:

$$|\mathfrak{N}^k V| = |\mathfrak{T}^k V| = |\mathfrak{L}^k V| = |\mathfrak{A}^k V| \tag{C.11}$$

Obviously, we have:

$$0 < |V| = |V \cup \Theta| \leq |V \cup I| \leq |V \cup L| \leq |V \cup D| < \infty \tag{C.12}$$

Finally, the following inequality holds, such that:

$$0 < \frac{|\mathfrak{A}^k V|}{|V \cup D|} \leq \frac{|\mathfrak{L}^k V|}{|V \cup L|} \leq \frac{|\mathfrak{T}^k V|}{|V \cup I|} \leq \frac{|\mathfrak{N}^k V|}{|V \cup \Theta|} \tag{C.13}$$

***Theorem 3.2*** **(Bounded Iterative Operators)** For all operators analyzed in this study, if $k > K, \forall k \in \mathbb{N}$, these iterative operators can be considered as consistent operator. If we have: $\|\hat{V}\| \leq \varepsilon$, it means: $\|\mathfrak{N}^k V - \mathfrak{A}^k V\| \to 0$, $\|\mathfrak{L}^k V - \mathfrak{A}^k V\| \to 0$ and $\|\mathfrak{T}^k V - \mathfrak{A}^k V\| \to 0$.

***Proof***: We denote: $\mathfrak{A}, \mathfrak{L}, \mathfrak{T}, \mathfrak{N} \in \mathfrak{C}: \mathbb{R}^{s \times t} \to \mathbb{R}^{s \times t}$

If $k > K$, For any operator belongs to $\mathfrak{C}$ can be considered as a contraction operator, according to Lemma 3.1, and we have:

$$\begin{aligned}\|\mathfrak{N}^k V - \mathfrak{A}^k V\| &= \|\mathfrak{N}^k V - \mathfrak{N}^k \hat{V} + \mathfrak{N}^k \hat{V} - \mathfrak{A}^k \hat{V} + \mathfrak{A}^k \hat{V} - \mathfrak{A}^k V\| \\ &\leq \|\mathfrak{N}^k V - \mathfrak{N}^k \hat{V}\| + \|\mathfrak{N}^k \hat{V} - \mathfrak{A}^k \hat{V}\| + \|\mathfrak{A}^k \hat{V} - \mathfrak{A}^k V\|\end{aligned} \tag{C.14}$$

According to Lemma 3.1, and we have:

$$\begin{aligned}\|\mathfrak{N}^k V - \mathfrak{N}^k \hat{V}\| &\leq \frac{\varepsilon}{3} \\ \|\mathfrak{A}^k \hat{V} - \mathfrak{A}^k V\| &\leq \frac{\varepsilon}{3}\end{aligned} \tag{C.15}$$

Considering the inequality:

$$\|\mathfrak{N}^k \hat{V} - \mathfrak{A}^k \hat{V}\| \leq \|\mathfrak{N}^k - \mathfrak{A}^k\| \cdot \|\hat{V}\| \tag{C.16}$$

Obviously, $\|\mathfrak{N}^k - \mathfrak{A}^k\| \leq M$, and we choose $\|\hat{V}\| \leq \frac{\varepsilon}{3M}$; it implies:

$$\|\mathfrak{N}^k \hat{V} - \mathfrak{A}^k \hat{V}\| \leq \frac{\varepsilon}{3} \tag{C.17}$$

And we have:

$$\|\mathfrak{N}^k V - \mathfrak{A}^k V\| \leq \varepsilon \tag{C.18}$$

# Appendix D

In this appendix, we provide the details to implement the DELMAR. The Algorithm 1 is the core algorithm to implement DELMAR. And Algorithm 2 describe the vital technique RRO used in DELMAR. The Algorithm 3 and 4 are important techniques employed to estimate the rank of feature matrices. Algorithm 5 describes the implementation of matrix back propagation (MBP).

---

**Algorithm 1 (Core Algorithm):** Deep Matrix Fitting (DELMAR)

---

***Input***: $SG \in \mathbb{R}^{t \times m}$, $SG$ is the input signal matrix;

Set $\beta > 1$ as the penalty parameter and $\gamma$ as the step-length

Randomly initialize $\{X_k\}_{k=1}^{M}$, $\{Y_k\}_{k=1}^{M}$ and $\{Z_k\}_{k=1}^{M}$ ; $M$ is large enough;

Set t as the initial estimated rank of $X_1$ and $Y_1$ and layer $k$ as 1.

   ***while*** rank > 1

      update $X_k$ using Eq. (3-1);

      update $Y_k$ using Eq. (3-2);

      update $Z_k$ using Eq. (3-3);

      $Z_k \leftarrow \mathcal{S}(Z_k)$;

      update multiplier $e_k$ using Eq. (3-4);

      use **Algorithm 2** to estimate rank of $Y_k$;

      $k \leftarrow k + 1$;

  ***end while***

  $K \leftarrow k;$

  Use **Algorithm 5** to perform matrix back propagation;

***Output***: $\{X_i\}_{i=1}^{K} \in \mathbb{R}^{n \times m}$, $\{Y_i\}_{i=1}^{K} \in \mathbb{R}^{n \times m}$ and $\{Z_i\}_{i=1}^{K} \in \mathbb{R}^{n \times m}$ ;

**Algorithm 2:** Rank Reduction Operator (RRO)

---

***Input:*** $Y_k \in \mathbb{R}^{n \times m}$, $Y_k$ is the feature matrix;

$YR_k \leftarrow QR(Y_k)$;

$minRank \leftarrow \min(1, size(Y_k))$;

$estRank \leftarrow minRank - 1$;

$diagXR_k \leftarrow abs(diag(YR_k))$;

using **Algorithm 3** calculate the weighted difference of $diagXR_k$;

set weighted difference of $diagXR_k$ as *wd*;

$[rankMax1, posMax1] \leftarrow \max(wd)$;

using **Algorithm 4** to calculate the weighted ratio of $diagXR_k$;

set weighted ratio of $diagXR_k$ as *wr*;

$[rankMax2, posMax2] \leftarrow \max(wr)$;

***if*** *rankMax1* is equal to *1*

$\quad estRank \leftarrow posMax1$;

***end if***

*valWR*←find( *wr>rankMax2* );

***if*** number (valWR) is equal to 1

$\quad estRank \leftarrow posMax2$;

***end if***

***if*** $\max(posMax1, posMax2)$ is equal to $\min(size((Y_k)))$

$\quad estRank \leftarrow rank - 1$

***else***

$\quad estRank \leftarrow \max(posMax1, posMax2)$;

***end if***

***Output:*** $estRank$

**Algorithm 3:** Weighted Difference (WD)

**Input:** $Vec \in \mathbb{R}^{1 \times m}$, $Vec$ is a vector;

$cumSum \leftarrow$ Calculate cumulative sum of $Vec$;

$diffVec \leftarrow$ Calculate differences between adjacent elements of $Vec$ ;

$WD \leftarrow$ -$diffVec$ ./ $cumSum$;

**Output:** WD

---

**Algorithm 4:** Weighted Ratio (WR)

**Input:** $Vec \in \mathbb{R}^{1 \times m}$, $Vec$ is a vector;

$L \leftarrow$ Calculate length of $Vec$;

$ratioVec \leftarrow Vec(1{:}L{-}1)$ ./ $Vec(2{:}L)$;

$WR \leftarrow (L{-}2)*ratioVec$ ./ sum $(ratioVec)$;

**Output:** WR

---

**Algorithm 5: Matrix Back Propagation (MBP)**

**Input:** $\{X_i\}_{i=1}^{M} \in \mathbb{R}^{T \times S}$, $\{Y_i\}_{i=1}^{M} \in \mathbb{R}^{T \times S}$, set $Z_i \leftarrow \prod_{j=1}^{i} X_i Y_i$;

*for* $k = T$ to 1

$\quad \psi \leftarrow \prod_{i=1}^{k-1} X_i$;

$\quad D_k \leftarrow \psi^{\dagger} Z_k \hat{Y}_k^{\dagger}$;

$\quad$ *if* $k < M$

$\quad\quad \hat{\alpha}_k \leftarrow \alpha_M$

$\quad$ *else*

$\quad\quad \hat{\alpha}_k \leftarrow D_{k+1} \hat{\alpha}_{k+1}$,

$\quad$ *end if*

$\quad \hat{Y}_k^+ \oplus \hat{Y}_k^- \leftarrow \hat{Y}_k$ ;

$\quad \hat{Y}_k^+ \leftarrow Y_k^+ \odot \sqrt{\dfrac{[\psi^T Z_k]^+ + [\psi^T \psi]^- \hat{Y}_k^+}{[\psi^T Z_k]^- + [\psi^T \psi]^+ \hat{Y}_k^+}}$;

$$|\tilde{Y}_k^-| \leftarrow |\hat{Y}_k^-| \odot \sqrt{\frac{[\psi^T Z_k]^+ + [\psi^T \psi]^- |\tilde{Y}_k^-|}{[\psi^T Z_k]^- + [\psi^T \psi]^+ |\tilde{Y}_k^-|}};$$

$$\tilde{Y}_k^- \leftarrow -|Y_k^-|;$$

$$Y_k \leftarrow \tilde{Y}_k^+ \oplus \tilde{Y}_k^-;$$

**end for**

**Output:** $\{X_i\}_{i=1}^K \in \mathbb{R}^{n \times m}, \{Y_i\}_{i=1}^K \in \mathbb{R}^{n \times m}$;

# Appendix E

The following figure provides all twelve reconstruction of FCs using DELMAR and other peer 3 methods, qualitatively compared with 12 ground truth templates.

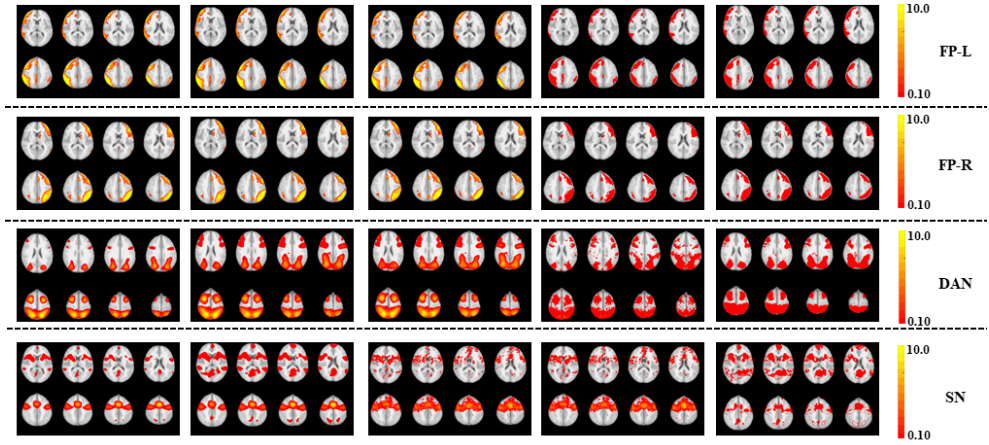

Figure S1. A presentation of all reconstructed 1st layer BCNs via DELMAR and other 3 peer method.